\documentclass[letterpaper, 10 pt, conference]{ieeeconf}  

\IEEEoverridecommandlockouts                              

\makeatletter
\let\NAT@parse\undefined
\makeatother
\usepackage{graphics} 
\usepackage{graphicx}
\usepackage{grffile} 
\usepackage{amsmath} 
\usepackage{amssymb}  
\usepackage{color}
\usepackage{mathtools} 
\usepackage{booktabs,ragged2e}
\usepackage{breqn}
\usepackage{textgreek}
\usepackage{arydshln}
\usepackage[flushleft]{threeparttable}
\usepackage{tabularx,booktabs}
\usepackage{graphics}
\usepackage{amsmath}
\usepackage{gensymb}
\usepackage[normalem]{ulem} 
\usepackage{pifont}
\usepackage{booktabs}
\usepackage{multirow}

\usepackage{lineno}
\usepackage{cite}
\usepackage{bm}
\usepackage{url}

\usepackage[noend]{algpseudocode}
\usepackage{algorithm}
\usepackage{makecell}
\usepackage[square, comma, numbers,sort&compress]{natbib}
\usepackage{multirow}
\let\labelindent\relax
\usepackage{enumitem}
\usepackage[table,dvipsnames]{xcolor}
\usepackage[pagebackref=true,breaklinks=true,colorlinks,bookmarks=false]{hyperref}
\usepackage{comment}
\usepackage{svg}
\hypersetup{
linkcolor=BrickRed
,citecolor=Green
,filecolor=Mulberry
,urlcolor=NavyBlue
,menucolor=BrickRed
,runcolor=Mulberry
,linkbordercolor=BrickRed
,citebordercolor=Green
,filebordercolor=Mulberry
,urlbordercolor=NavyBlue
,menubordercolor=BrickRed
,runbordercolor=Mulberry
}

\usepackage[colorinlistoftodos]{todonotes}
\usepackage[normalem]{ulem} 
\newcommand{\uksang}[1]{\todo[inline, color=blue!20]{Uksang: #1}}
\renewcommand{\uksang}[1]{}

\usepackage{float}
\usepackage{xcolor}
\usepackage{soul}
\usepackage{balance}

\title{\LARGE \bf
One-shot Video Imitation via Parameterized Symbolic Abstraction Graphs
}
\author{\small Jianren Wang \textsuperscript{1 } \quad \small Kangni Liu \textsuperscript{2 } \quad \small Dingkun Guo \textsuperscript{3 } \quad \small Zhou Xian \textsuperscript{4 } \quad \small Christopher G. Atkeson \textsuperscript{5 }
\thanks{$^{1,2,3,4,5}$Robotics Institute, Carnegie Mellon University, Pittsburgh, USA
{\tt\footnotesize \{jianrenw, kangnil, dingkung, xianz1, cga\}@andrew.cmu.edu}}}

\begin{document}

\maketitle
\thispagestyle{empty}
\pagestyle{empty}
\renewcommand{\baselinestretch}{0.979} 

\newcommand{\website}{\href{https://jianrenw.com/PSAG/}{https://www.jianrenw.com/PSAG/}}

\begin{abstract}
Learning to manipulate dynamic and deformable objects from a single demonstration video holds great promise in terms of scalability. Previous approaches have predominantly focused on either replaying object relationships or actor trajectories. The former often struggles to generalize across diverse tasks, while the latter suffers from data inefficiency. Moreover, both methodologies encounter challenges in capturing invisible physical attributes, such as forces. In this paper, we propose to interpret video demonstrations through Parameterized Symbolic Abstraction Graphs (PSAG), where nodes represent objects and edges denote relationships between objects. We further ground geometric constraints through simulation to estimate non-geometric, visually imperceptible attributes. The augmented PSAG is then applied in real robot experiments. Our approach has been validated across a range of tasks, such as Cutting Avocado, Cutting Vegetable, Pouring Liquid, Rolling Dough, and Slicing Pizza. We demonstrate successful generalization to novel objects with distinct visual and physical properties. For visualizations of the learned policies please check: \website
\end{abstract}

\renewcommand{\baselinestretch}{0.979} 
\section{Introduction}

Humans can learn to manipulate dynamic and deformable objects by simply watching one \textit{single} demonstration video. It is desired for robots to acquire similar capabilities and learn from the massive amount of videos covering numerous skills available on the internet, which could potentially bring a "ChatGPT moment" to the field of robotics and make a significant step forward in robot learning and autonomy. However, despite considerable progress in artificial intelligence and robotics in recent years, the development of robot systems that can match human-level capabilities of learning from video demonstrations remains elusive.
Why is this difficult? We identify three primary challenges in learning from few-shot video demonstrations. First, the extraction of the demonstrator's intentions and actions from videos is complicated by the presence of extraneous information in videos, making it difficult to isolate relevant cues. Second, the task of translating observed intentions into adaptable skills—skills that robots can apply on a wide range of objects and environments, each with its own set of visual and physical characteristics—presents a significant challenge. This difficulty stems from the large variability in visual and physical attributes across different objects and environments. Third, developing efficient learning algorithms that enable robots to acquire skills from a limited number of demonstrations, is also intrinsically a difficult task.

Two primary approaches to learning from demonstration have been explored in the past.
One seminal approach developed nearly half a century ago
focused on interpreting the demonstrator's intentions as object relationships and replicating those relationships~\cite{winston1970learning}. Subsequent research in this domain focuses on utilizing perceived abstract relationships as objectives for planning techniques, including Task-level Planning~\cite{lozano1989task} and Task and Motion Planning~\cite{garrett2021integrated}. Notably, these approaches often overlook detailed metrics of objects and their relationships, and in many instances, the precise motions of actors are de-emphasized and sometimes not even measured. While these methods efficiently eliminate most irrelevant information and are adept at learning from a single demonstration, they often rely on strong human priors, incorporating hardcoded elements, which poses a challenge in addressing the first challenge~\cite{wen2022you}.

Another avenue in learning from demonstration involves direct robot teaching, wherein desired positions and trajectories are demonstrated to a robot through teleoperation~\cite{fu2024mobile}, motion capture technology~\cite{AMASS:ICCV:2019}, or computer vision~\cite{whirl}. The robot then learns to predict actions that align with the demonstration data~\cite{ross2011reduction, wang2023manipulate, chi2023diffusionpolicy}. These approaches assume that, through training on a diverse dataset, the system will implicitly comprehend the demonstrators' intentions. However, this assumption is not always accurate and can be highly inefficient, leading to a failure in addressing the third challenge.

More importantly, both approaches fail to address the second challenge, as most physical characteristics other than movement are challenging to observe from videos. One purpose of this paper is to make the point that behavior is more than just replaying object relationships or trajectories interpolated from a set of observed trajectories. This is especially true in tasks where the exact values of forces matter, such as in dynamic tasks~\cite{atkeson1997robot}, and tasks involving deformation, separation, and combination of materials. 

In contrast to the aforementioned approaches, our proposal involves interpreting video demonstrations using Parameterized Symbolic Abstraction Graphs (PSAG). These graphs consist of nodes and edges as abstractions, which are parameterized by their geometric and non-geometric attributes. In this framework, each node can represent a rigid or deformable object, incorporating attributes that capture the six degrees of freedom (6DOF) of the object pose. The edge information defines relationships between objects, encompassing aspects like contact, and is parameterized according to the contact region and forces acting between them.

To build the PSAG, we begin by utilizing off-the-shelf detectors, depth estimators, and optical flow estimators to evaluate object poses and relationships. Subsequently, we learn to simulate the demonstration with this geometric information (\textit{e.g.} positions, contact points) to calculate forces. This enables us to parameterize non-geometric imperceptible attributes (e.g., forces), laying the groundwork for transferring the skill to the real world.

Our proposed method helps the agent disregard irrelevant information and effectively learn from a \textbf{single} video. We validate the efficacy of our approach through experiments conducted on five challenging tasks: \textit{Cutting Avocado}, \textit{Cutting Vegetable}, \textit{Pouring Liquid}, \textit{Rolling Dough} and \textit{Slicing Pizza}. Notably, the test environments differ substantially from the learning environments, encompassing variations in geometry, appearance, and physics. 

To summarize, our contribution includes:

\begin{itemize}
    \item Proposing to interpret video demonstrations as parameterized symbolic abstraction graphs (PSAG)
    \item Proposing to learn simulations from demonstrations with minimal human input, enabling the addition of non-geometric temporally parameterized visually imperceptible attributes to the edges.
    \item Demonstrating the efficacy of our approach in performing dynamics and deformable manipulation tasks with generalizability.
\end{itemize}

\renewcommand{\baselinestretch}{0.979} 
\section{Related Work}

\paragraph{Symbolic Visual Reasoning}

Symbolic methods have shown good data efficiency and generalization capabilities across various computer vision tasks, ranging from visual question answering~\cite{andreas2016learning, andreas2016neural, hudson2019learning} to image captioning~\cite{anderson2016spice, johnson2015image, krishna2017visual}. Recent advances have extended these methods to 3D object relationship reasoning~\cite{hsu2023ns3d, armeni20193d, kim20193}, object physics modeling~\cite{gupta2010blocks, chen2021grounding, ates2020craft, melnik2023benchmarks, chen2022comphy, ding2021dynamic, wu2017learning}, and video action understanding~\cite{mavroudi2020representation, rai2021home, lu2016visual}.

In this work, we extend the ability of previous approaches to perceive physics and exploit physical constraints from a given video~\cite{gupta2010blocks, zhang2023slomo} to guide robots in performing tasks involving diverse objects and environments. And in contrast to approaches that emphasize reasoning about intervention~\cite{li2021intervention}, which predominantly revolve around predicting outcomes, our work focuses on predicting how to achieve the same desired outcome when presented with a new set of objects.

\paragraph{Learning from Human Video}

Learning from Demonstration (LfD)~\cite{argall2009survey, billard2008survey, schaal1999imitation} focuses on learning from expert demonstrations~\cite{chi2023diffusionpolicy, wang2023manipulate, fu2024mobile}. These approaches often have limitations as they rely on a large number of expert demonstrations and assume a shared observation and action space between the imitator and demonstrator. These constraints significantly restrict the potential for effective learning from videos. Instead of learning from robot demonstrations, an alternative approach involves learning from human demonstrations, which are easy and cost-effective to collect. However, the challenge lies in the absence of ground truth actions. One direct method is to imitate human motion~\cite{bharadhwaj2023zero, wang2023mimicplay, whirl, xiong2021learning} or object motion~\cite{atkeson1997robot, wen2022you, zhu2024vision}. Often, this line of work focuses on replicating low-level actions of the demonstration rather than developing more generalizable abstractions. In the pursuit of learning higher-level abstractions related to manipulation, some studies attempt to predict visual affordances, indicating where to interact in an image and providing local information on how to interact~\cite{goyal2022human, liu2022joint}. While these approaches can serve as effective initializations for a robotic policy, they are not standalone solutions for task completion. Typically, they are employed in conjunction with online learning, necessitating several hours of deployment-time training and robot data~\cite{bahl2023affordances}. Importantly, these affordances fall short for complex tasks, such as cutting, where crucial information like force cannot be observed from the video, limiting the applicability to tasks beyond simple pick-and-place tasks. In contrast to previous approaches, our proposed method grounds the object relationships in simulation. This allows the robot to uncover unseen information from videos and thus facilitates learning from a diverse array of tasks.

\paragraph{Deformable Object Manipulation}Deformable object manipulation poses a longstanding challenge in robotics. Prior research has primarily concentrated on various tasks such as rope manipulation~\cite{nair2017combining}, pouring liquid~\cite{li20223d}, and cloth manipulation~\cite{ha2022flingbot}. Additionally, manipulating elastoplastic objects, such as deforming them through grasping~\cite{shi2024robocraft}, rolling~\cite{lin2023planning}, or cutting~\cite{heiden2021disect}, has been explored in other studies. Instead of individually learning each skill through model-free reinforcement learning or model-based planning, our paper focuses on learning multiple skills from multiple videos.

\paragraph{Real-to-sim-to-real Transfer} Prior work has employed 3D reconstruction techniques to create realistic scene representations for improving manipulation~\cite{torne2024reconciling}, navigation~\cite{deitke2023phone2proc}, and locomotion~\cite{byravan2023nerf2real}. However, these approaches 1) require thousands of views for accurate scene reconstruction and 2) depend on human demonstration collection or reward design for reinforcement learning~\cite{torne2024reconciling}. In contrast, our approach 1) does not rely on detailed scene reconstructions and 2) eliminates the need for expert-designed rewards.

\renewcommand{\baselinestretch}{0.979} 
\section{Methods}

\begin{figure*}
    \centering 
    \includegraphics[width=0.85\linewidth]{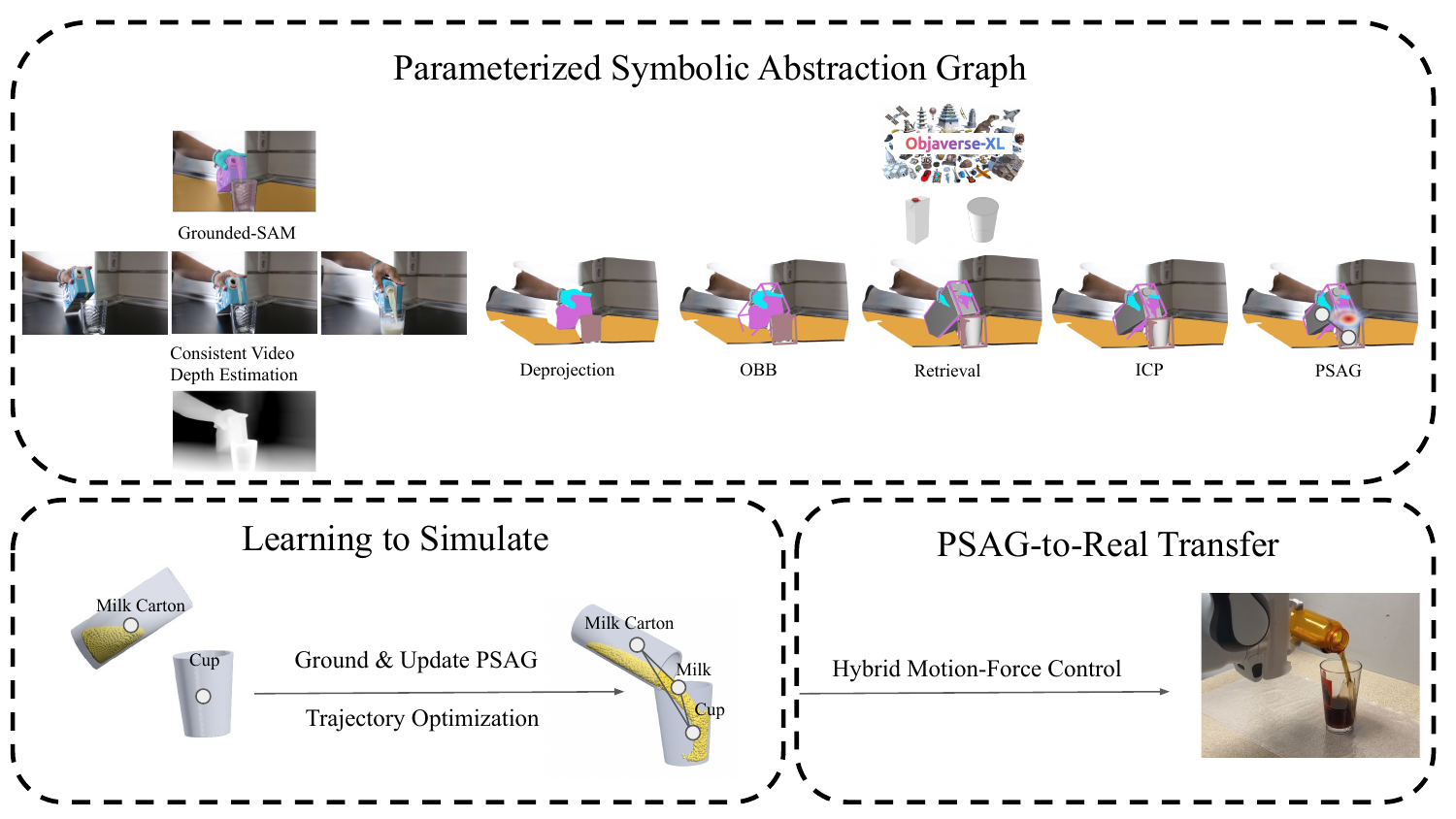} 
        \caption{Overview of our pipeline for learning from videos. (a) Building Parameterized Symbolic Abstraction Graphs (PSAG): PSAGs are generated by instance segmentation, consistent video depth estimation, and object relationship calculation. (b) Learning to Simulate: Constructing digital twin to ground geometric constraints via trajectory optimization (c) PSAG-to-Real Transfer using hybrid motion-force control.}
    \label{fig:pipeline}
\end{figure*}

In this section, we first outline the process of constructing parameterized symbolic abstraction graphs (PSAG) from videos using off-the-shelf tools (Sec.~\ref{sec:scene}). Following that, we describe the process of learning simulations from PSAG and adding non-geometric visually imperceptible attributes to the edges. (Sec.~\ref{sec:twin}). We then explain the method for generating a real-world executable program from a PSAG.(Sec.~\ref{sec:sim2real}) (Fig.\ref{fig:pipeline}).

\subsection{Building PSAGs}
\label{sec:scene}

To construct parameterized symbolic abstraction graphs from videos, our approach consists of three key steps. First, we employ computer vision techniques to extract depth information, perform instance segmentation, and calculate optical flow, which enable us to reconstruct a semantic point cloud for each frame. Next, we fine-tune the depth estimator by incorporating 3D geometric constraints to achieve temporal consistency in video depth estimation. Finally, we retain only objects of interest and calculate their attributes and relationships with other objects, facilitating the construction of the PSAG. We now introduce each module in detail.

\paragraph{Semantic Point Cloud Reconstruction}

We commence by estimating the monocular depth using ZoeDepth~\cite{bhat2023zoedepth}, which provides metric-scale depth information for each RGB frame. It is crucial to note that the depth information across adjacent frames lacks coherence, an issue addressed in subsequent paragraphs. Subsequently, we employ Grounding DINO~\cite{liu2023grounding} for object detection and Segment Anything (SAM)\cite{Kirillov_2023_ICCV} for instance segmentation. We leverage GMFlow\cite{xu2022gmflow} for optical flow estimation, which facilitates object tracking across frames. This process enables us to construct a semantic point cloud that encapsulates both spatial and semantic details.

\paragraph{Consistent Video Depth Estimation}

We employ Consistent Video Depth Estimation (CVDE)~\cite{luo2020consistent} to produce temporally coherent and geometrically consistent depth maps throughout the entire video. CVDE fine-tunes the pre-trained single-image depth estimation model~\cite{bhat2023zoedepth} to minimize geometric inconsistency errors across multiple frames specific to the given video. Following the fine-tuning stage, the final depth estimation results for the video are computed using the fine-tuned model.

\paragraph{PSAG Generation}

To generate the PSAG, our first step is to filter out irrelevant objects. We specifically identify objects that interact with the hand as objects of interest. This concept of interaction can be hierarchically propagated. For example, objects directly interacting with the hand are classified as the first level of interaction, and objects interacting with the first-level objects are classified as the second level of interaction, and so on. In our current implementation we preserve objects with up to three levels of interaction.

Next, we generate a graph representation to capture attribute changes and relationship dynamics among the objects of interest. Given that reconstructing dynamic scenes from single videos remains an open problem with no comprehensive solution, we propose a retrieval-based approach for estimating 6DOF object poses. Initially, we compute an oriented bounding box (OBB) around each object using the method from~\cite{o1985finding}. Next, employing Pointnet++ \cite{qi2017pointnet++}, we retrieve the nearest neighbor from a subset of Objaverse-XL \cite{deitke2023objaverse}. We then resize and orient the retrieved object to fit the OBB of each corresponding object. Due to potential occlusion, the estimated OBBs may not tightly bound the objects. To address this, we further refine the object poses using iterative closest points (ICP)~\cite{beslpj1992amethod}, which are used are node attributes. Additionally, we incorporate edge information to indicate whether two objects are in contact with each other. This involves calculating the Chamfer distance between their respective point clouds. If the minimum distance falls below a predefined threshold, we determine that the two objects are in contact. Additionally, we capture the closest points of each pair of objects in contact. To enhance the representation, we apply Gaussian filters to smooth the contact region, and these smoothed regions are then utilized for optimization. Through these processes, we construct a PSAG for a given video. In this graph, each node corresponds to an object of interest, and each edge denotes the relationship between them. It's worth noting that the current edge attributes only include geometric information (\textit{e.g.} contact regions). We will now elaborate on how to incorporate non-geometric visually imperceptible attributes into the edges.

\subsection{Learning to Simulate}
\label{sec:twin}

With the provided PSAG, we propose to learn simulations for estimating non-geometric attributes. Considering that most general tasks from videos involve the interaction of both rigid and deformable bodies, we advocate the utilization of the Moving Least Squares Material Point Method (MLS-MPM)\cite{hu2018moving}, as per the approach outlined in\cite{xian2023fluidlab, xu2023roboninja}. 

To ground the object relationships over the entire video, we initially decompose each task into multiple subtasks based on changes in constraints, such as the establishment and breaking of contacts. For each subtask, we utilize the initial object relationships of the subsequent subtask as optimization goals. We model this process within a Markov Decision Process (MDP) defined by a set of states $s\in\mathcal{S}$, actions $a\in\mathcal{A}$, and a deterministic, differentiable transition dynamics $s_{t+1} = p(s_t, a_t)$, where $t$ denotes discrete time, and states are composed of different objects $s_t = \{s_t^i\}_{i=1,...,n}$. For any pair of objects $(s_t^i, s_t^j)$, the geometric constraints can either exist (in contact) or not (not in contact), and a cost function is denoted as $C(s_t^i, s_t^j)$. For any reference state (6DOF poses of objects) $\hat{s_t}$, a distance function is denoted as $D(s_t, \hat{s_t})$. Following Wen \textit{et al.}~\cite{wen2022you}, object poses are expressed in the receptacle’s coordinate frame (\textit{e.g.}, milk carton’s pose relative to the cup in the pouring task), which allows generalization to new scene configurations regardless of absolute poses. We also adopted the Non-uniform Normalized Object Coordinate Space (NUNOCS)~\cite{wen2022catgrasp} for category-level trajectory projection, which enables generalization to arbitrary, unknown instances. The objective is to determine a trajectory that minimizes the total loss $L$. Following~\cite{lin2022diffskill}, we use gradient-based trajectory optimization to solve for an open-loop action sequence:

\begin{align*}
    \operatorname*{argmin}_{a_0, ..., a_{T-1}} L(a_0, ..., a_{T-1}) = \operatorname*{argmin}_{a_0, ..., a_{T-1}} \lambda_1 \times \sum_{t=1}^{T} \sum_{i,j} C(s_t^i, s_t^j) \\
    + \lambda_2 \times \sum_{t=1}^{T} D(s_t, \hat{s_t}) + \lambda_3 \times \sum_{t=1}^{T} E(a_t) \tag{1} \label{eq:1}
\end{align*} where $s_{t+1} = p(s_t, a_t)$.

$C(s_T^i, s_T^j)$ represents the KL divergence~\cite{kullback1951information} between contacting distributions if there are geometric constraints between $(s_t^i, s_t^j)$; otherwise, it is 0. We have additionally formulated a cost function for scenarios involving the separation of one object into two parts, such as cutting. In this context, the cost is defined as the minimum distance between each pair of separated parts. The action $a_t$ encompasses both the translation velocity and angular velocity of each object, and $E(a_t)$ denotes the energy associated with executing action $a_t$. Additionally, $\lambda_1, \lambda_2, \lambda_3$ are weighting parameters. 

We address Equation~\ref{eq:1} by iteratively updating the action sequence with $\nabla L_{a_t}$, where $t$ ranges from 0 to $T$, employing an Adam optimizer~\cite{kingma2014adam} initialized with reference trajectories. The modified trajectories are then utilized to update the geometric attributes of the PSAG. Furthermore, at each timestep, we compute the force $f_t$ and torque $\tau_t$ observations, which are incorporated into the PSAG edge attributes. This augmented PSAG enables the implementation of a hybrid motion-force controller for real-world applications.

\subsection{PSAG-to-Real Transfer}
\label{sec:sim2real}

Instead of transferring an end-to-end policy that directly operates in real environments based on visual inputs~\cite{zhuang2023parkour, chen2023visual}, our approach involves transferring abstract poses and forces. Notably, the PSAG in Sec~\ref{sec:twin} is represented in the receptacle's coordinate frame and Non-uniform Normalized Object Coordinate Space (NUNOCS) to enable category-level generalization. To adapt them to the real world, we must update the PSAG with the current environment, transitioning from the receptacle's coordinate frame and NUNOCS to the real-world frame at the actual scale.

\begin{figure*}
    \centering 
    \includegraphics[width=0.85\linewidth]{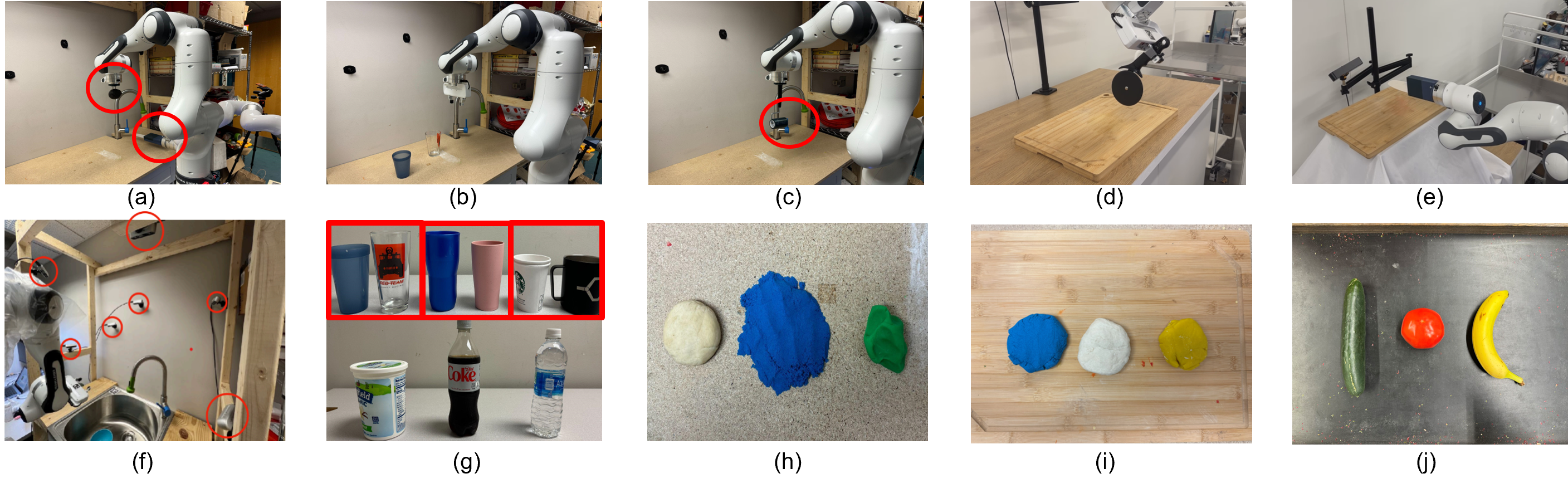} 
        \caption{Experiment Settings: (a) Robot arm with an avocado holder and another arm with a knife and force sensor.(b) Robot arm and cups for the pouring task. (c) Rolling pin mounted on the robot arm for rolling dough. (d) Slicer affixed to the robot arm for slicing pizza. (e) Knife mounted on the robot arm for cutting vegetables. (f) Multi-camera system. (g) Cups, yogurt, Coke, and water for the pouring task. (h, i) Dough, play sand, and play dough for the rolling dough and slicing pizza experiments. (j) Cucumber, tomato, and banana for the cutting vegetables task.}
    \label{fig:setting}
\end{figure*}

The initial step involves transitioning from the receptacle's frame to the world frame, requiring knowledge of object poses and shapes in the real world. We constructed a multi-camera system as depicted in Fig.~\ref{fig:setting}(f). Utilizing eight RealSense D435 depth cameras calibrated with Multical~\cite{oliver2023} using an AprilTag~\cite{edwin2011} board, we deproject the depth and color images into a point cloud. By consolidating information from all eight cameras, we obtain a comprehensive spatial understanding. Following segmentation, we crop the point cloud of each object and fit an object to the position as mentioned in Sec~\ref{sec:scene}, providing us with the position and orientation of each object.

The second step is to adopt NUNOCS to capture the geometric variation across all instances, which is not direct. Consider milk pouring as an example. If the commonly selected center-of-mass (CoM) is used as the canonical coordinate frame origin when aligning a novel object instance to the demonstrated one, it may collide with or float away from the receptacle. To address this, we initialize the trajectory from the processed PSAG, where the origin of the category-level canonical coordinate system is the CoM. Then we redo the process described in Sec~\ref{sec:twin} again with the test-time objects, but this time, everything is represented in the world frame with its real size. After this, we obtain the optimal trajectory in the real-world frame at the actual object scale.

Finally, we transfer optimal trajectories from the simulation to a real robot using a hybrid motion-force controller. The transfer mechanism is described by Equation~\eqref{eq:2}:
\begin{align*}
p_t^r = k_1 \times p_t^s + k_2 \times (f_t^r - f_t^s) \tag{2} \label{eq:2}
\end{align*}
Here, $p_t^r$, $f_t^r$ denote the positions and forces of the real robot, while $p_t^s$, $f_t^s$ represent the positions and forces obtained from simulation. The parameters $k_1$ and $k_2$ are compliance parameters. The robot is position-controlled in Cartesian space to achieve both the target position and the desired force. We then employ inverse kinematics to convert Cartesian coordinates to joint space, where the robot is controlled by a Proportional-Derivative (PD) controller.

\renewcommand{\baselinestretch}{0.979} 
\section{Experiments}

In this section, we present experiments conducted in both simulated and real-world environments, followed by results and ablation studies.

\subsection{Experimental Setup}

\paragraph{Tasks} Our experiments were designed to evaluate the performance of our approach on five tasks involving deformable objects: \textit{Cutting Avocado}, \textit{Cutting Vegetable}, \textit{Pouring Liquid}, \textit{Rolling Dough} and \textit{Slicing Pizza}. Deformable objects are ubiquitous in kitchen settings, an important environment for future robotic applications. We selected these tasks to evaluate the model's generalizability and adaptability, which are important for its success in real-world scenarios. 


\paragraph{Evaluation} We devised task-specific metrics for qualitative evaluation over 20 trials for each task. For cutting avocados, success is cleanly cutting around the core without cutting into it. For vegetables, success is making a smooth planar cut without generating excessive force on the cutting board. For pouring liquids, success is pouring Coca-Cola/water/yogurt into the cup without spilling. For rolling dough, success is maintaining contact and flattening play sand/play-dough/dough without separating pieces. For slicing pizza, success is dividing it into two parts without generating excessive force or tearing the pizza (Figure~\ref{fig:setting}).

\paragraph{Baselines and Ablations} We evaluate our method against four variants of existing approaches: You Only Demonstrate Once (YODO)\cite{wen2022you}, Trajectory Following (TF), Inverse Reinforcement Learning (IRL), and Interaction Warping (IW)\cite{biza2023one}. YODO and TF focus on replaying the trajectory observed in the video, while IRL optimizes object relationships without using trajectories. All baselines solely leverage features directly observable in the video. 


\begin{figure*}
    \centering 
    \includegraphics[width=0.86\linewidth]{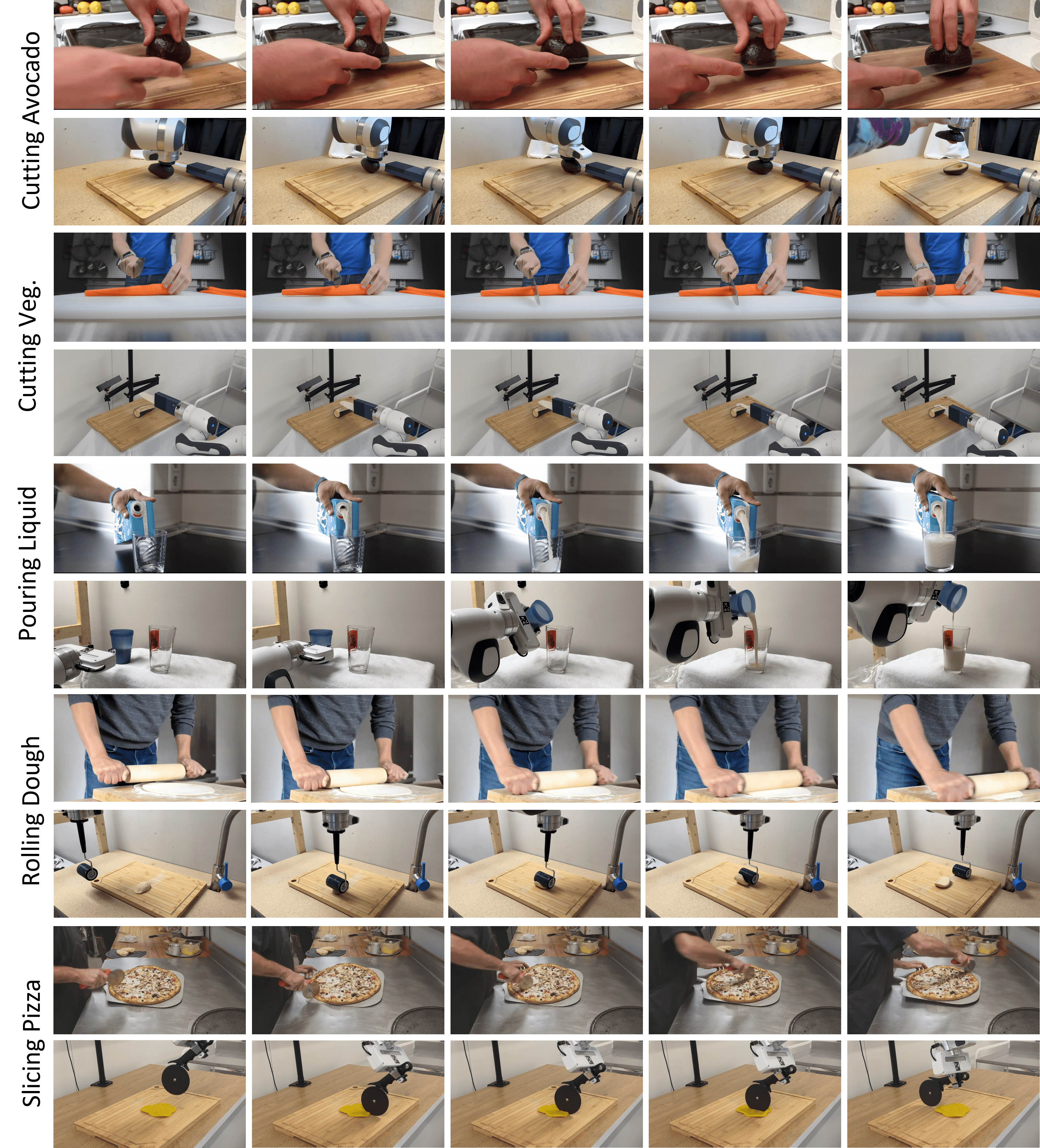} 
        \caption{For each task, we present the video demonstration (top) and the robot trajectories (bottom). Our proposed method allows the robot to perform challenging tasks such as cutting an avocado, cutting vegetables, pouring liquid, rolling dough, and slicing pizza from a single demonstration.}
    \label{fig:qualitative}
\end{figure*}



\begin{table*}[t]
\centering
\renewcommand{\arraystretch}{1.1}
\scalebox{0.82}{
\begin{tabular}{l|c|c|c|c|c}

Method & Cutting Avocado & Cutting Vegetable & Pouring Liquid & Rolling Dough & Slicing Pizza\\ \toprule
YODO & 10\% & 55\% & 15\% & 10\% & 50\%\\ 
TF & 5\% & 50\% & 5\% & 10\% & 40\%\\ 
IRL & 0\% & 0\% & 10\% & 0\% & 0\% \\
IW & 10\% & 65\% & 80\% & 25\% & 60\%\\ \midrule
Ours & 75\% & 75\% & 80\% & 70\% & 70\%                                                       
\\ \toprule
\end{tabular}
}
\caption{Quantitative Results: Each row represents the success rate of the five tasks. Rows 1-4 demonstrate the results of You Only Demonstrate Once, Trajectory Following, Inverse Reinforcement Learning, and Interaction Warping, which serve as baseline methods for comparison. Row 5 illustrates that our method consistently outperforms baseline methods by a large margin.}
\label{tab:result}
\end{table*}

As shown in Table~\ref{tab:result}, our method outperforms all baseline methods by a large margin. The three components—geometric constraints, reference trajectories, and force information—each contribute to learning a robust policy. Comparing YODO and TF with our method, the absence of geometric constraints in these baselines results in no reward mechanism promoting behaviors like rolling or pouring, leading to a lower success rate. Comparing IRL with our method shows that omitting reference trajectories and only encouraging geometric constraints leads to either aggressive trajectories (e.g., pouring) or getting trapped in a local minimum, preventing the acquisition of crucial rotational behavior necessary for task completion (e.g., cutting avocado), resulting in a low success rate. Comparing IW with our method, IW's results are heavily task-specific. For tasks where force is crucial (e.g., cutting avocado and rolling dough), our method performs significantly better than the baseline methods. This demonstrates that merely transferring observable information from the video to the robot program is not as effective as our approach. 

We also present the human demonstrations (from YouTube) and the sampled robot trajectories in Fig.~\ref{fig:qualitative}. Please refer to the supplementary video for more qualitative results.

\renewcommand{\baselinestretch}{0.979} 
\section{Limitations and Conclusions}


While our method enables one-shot video imitation across diverse tasks, it has three key limitations. First, errors in detection and depth estimation present major challenges, as precise results are difficult to achieve in real-world scenarios without extensive hyperparameter tuning. Second, tuning hyperparameters during learning, especially adjusting cost term weights based on policy performance, remains a complex issue, as seen in many trajectory optimization and reinforcement learning methods. Third, fine-tuning simulation properties, such as grid density or material viscosity, is another hurdle, with autonomous generalization across tasks still being difficult to achieve.

In conclusion, our work highlights that behavior extends beyond merely replicating object relationships or actor trajectories. We propose interpreting video demonstrations as Parameterized Symbolic Abstraction Graphs (PSAG), where nodes represent objects and edges signify relationships. By grounding geometric relationships in simulation and incorporating non-geometric, visually imperceptible attributes such as forces, our method effectively learns to manipulate diverse dynamics and deformable objects from a single video demonstration. This approach suggest ways to reduce both teleoperated robot demonstration data and expensive instrumentation of demonstrations to measure contact and forces and emphasizes learning from single human demonstrations.



\addtolength{\textheight}{-0.2cm}   


\bibliographystyle{ieeetr}

\bibliography{references.bib}

\newpage

\end{document}